\documentclass[letterpaper]{article} 
\usepackage{aaai2027}
\usepackage{times}  
\usepackage{helvet}  
\usepackage{courier}  
\usepackage[hyphens]{url}  
\usepackage{graphicx} 
\usepackage{natbib}  
\usepackage{caption} 
\usepackage{amsmath} 
\usepackage{amssymb} 
\usepackage{booktabs} 
\usepackage{array} 
\title{Gradual Cognitive Externalization: From Modeling Cognition to Constituting It}

\author{
    Zhimin Zhao
}
\affiliations{
    Software Analysis and Intelligence Lab (SAIL), Queen's University
}

\begin{document}

\maketitle

\begin{abstract}
Developers are publishing AI agent skills that replicate a colleague's communication style, encode a supervisor's mentoring heuristics, or preserve a person's behavioral repertoire beyond biological death. To explain why, we propose Gradual Cognitive Externalization (GCE), a framework arguing that ambient AI systems, through sustained causal coupling with users, transition from modeling cognitive functions to constituting part of users' cognitive architectures. GCE adopts an explicit functionalist commitment: cognitive functions are individuated by their causal-functional roles, not by substrate. The framework rests on the behavioral manifold hypothesis and a central falsifiable assumption, the \emph{no behaviorally invisible residual} (NBIR) hypothesis: for any cognitive function whose behavioral output lies on a learnable manifold, no behaviorally invisible component is necessary for that function's operation. We document evidence from deployed AI systems showing that externalization preconditions are already observable, formalize three criteria separating cognitive integration from tool use (bidirectional adaptation, functional equivalence, causal coupling), and derive five testable predictions with theory-constrained thresholds.
\end{abstract}

\section{Introduction}
\label{sec:intro}

In early 2026, thousands of developers began publishing AI agent ``skills'' that encode not just professional expertise but personal identity, including a colleague's communication style, a supervisor's mentoring heuristics, a parent's interaction patterns, one's own behavioral repertoire, and even an ``immortal skill'' designed to preserve a person beyond biological death. A curated catalog of such human-distillation skills already tracks dozens of projects across these categories~\cite{liu2025_awesome_distillation}. Simultaneously, SkillsBench demonstrated that externalized domain expertise measurably improves AI agent performance across diverse professional tasks~\cite{li2026skillsbench}. These are not isolated curiosities. Over two dozen AI platforms, including Claude Code, Cursor, and GitHub Copilot, now support a shared skill standard for loading domain knowledge on demand~\cite{agentskills2025}. Users are externalizing how they think, decide, and communicate into executable digital artifacts, and AI systems that absorb these artifacts become measurably more competent.

Most of these artifacts remain static persona descriptions rather than adaptive systems, as we detail below. Yet the impulse they reveal is significant: users are attempting to encode cognitive functions into executable digital form. We argue this impulse is the early stage of a deeper process in which ambient AI systems, through sustained causal coupling with users, transition from modeling cognitive functions to constituting part of users' cognitive architectures. We call this process Gradual Cognitive Externalization (GCE). The term ``externalization'' is used deliberately in two senses that GCE claims are connected: the weak sense of encoding cognitive patterns into digital artifacts (observable today), and the strong sense of those artifacts becoming genuine components of the user's cognitive architecture (the theoretical claim). GCE's central argument is precisely that the former transitions into the latter when three criteria are jointly satisfied (Section~\ref{sec:framework}). GCE claims cognitive function transfer, not consciousness transfer; because it involves no discrete transfer event, biological cognition remains unbroken throughout.

Traditional mind uploading requires complete neural mapping and faces unsolved conceptual problems~\cite{sandberg2008_whole, chalmers2010_singularity}. GCE sidesteps them by operating on behavioral data: ambient systems already learn planning patterns, communication styles, preference structures, and knowledge organization from sustained interaction (Section~\ref{sec:evidence}). What drives the transition from external model to cognitive component is not prediction fidelity alone but fidelity combined with bidirectional adaptation and causal coupling (Section~\ref{sec:framework}).

GCE builds on four foundations: (1) the behavioral manifold hypothesis (BMH), which holds that everyday human cognition occupies a low-dimensional learnable manifold in the high-dimensional space of possible behaviors; (2) extended mind theory, under which cognitive processes extend beyond biological boundaries through functionally integrated coupled systems~\cite{clark1998_extended, clark2025_extending}; (3) multiscale competency architecture, which shows that cognitive capabilities emerge through hierarchical information processing across diverse substrates~\cite{levin2022_technological}; and (4) functionalism, the thesis that cognitive functions are individuated by their causal-functional roles rather than by substrate~\cite{putnam1967_psychological}. Functionalism is non-negotiable for GCE: without it, no amount of behavioral equivalence would count as externalization.

GCE's commitment to behavioral observation aligns with Dennett's heterophenomenology~\cite{dennett1991_consciousness} and draws on the ``extended'' strand of 4E cognition~\cite{newen2018_4e}, while departing from the enacted strand's emphasis on sensorimotor coupling. The consciousness indicator framework proposed by~\citeauthor{butlin2023_consciousness}~\cite{butlin2023_consciousness} evaluates AI systems against neuroscientific theories of consciousness but focuses on spontaneous machine consciousness rather than externalization of human cognition. To our knowledge, no prior framework synthesizes extended mind theory, multiscale competency architecture, and the behavioral manifold hypothesis into a unified account of gradual cognitive externalization.

\section{Evidence for Externalization}
\label{sec:evidence}

Before formalizing GCE, we present evidence that its preconditions are already observable.

GCE predicts that cognitive functions with clear behavioral signatures will externalize first. Current systems confirm this across four domains. \textit{Temporal planning}: scheduling assistants (Google Calendar, Reclaim.ai\footnote{\url{https://reclaim.ai}}, Clockwise\footnote{\url{https://www.getclockwise.com}}) autonomously block time and optimize meetings based on learned individual patterns~\cite{cook2009_ambient}, acting \emph{within} the user's planning process. \textit{Communication style}: Gmail Smart Compose generates email completions that recipients accept without modification~\cite{chen2019_gmail}; Grammarly\footnote{\url{https://www.grammarly.com}} adapts to individual writing styles; GPT-based assistants produce messages evaluators cannot reliably distinguish from human-authored ones~\cite{clark2021_human_eval}. \textit{Preference structures}: collaborative filtering systems learn latent preferences that predict user selections with high accuracy, often surfacing choices users would not have identified through self-report~\cite{netflix2009_prize}, while computer-based personality models inferred from behavioral traces outperform human judges including close acquaintances~\cite{youyou2015_personality}. \textit{Knowledge organization}: Notion AI\footnote{\url{https://www.notion.com/product/ai}}, Mem\footnote{\url{https://get.mem.ai}}, and LangChain's LangMem SDK\cite{chase2022_langchain, langchain2025_ambient} maintain persistent, user-scoped memory that accumulates across sessions.

What distinguishes the most recent wave is the shift from implicit learning to explicit externalization. The open agent skills standard~\cite{agentskills2025} lets practitioners encode their decision heuristics into portable \texttt{SKILL.md} files that AI agents load on demand, functioning like procedural memory retrieval. Dedicated skill registries\footnote{\url{https://clawhub.ai}} and commercial marketplaces have already emerged around this standard. This is not passive data collection but deliberate cognitive externalization: practitioners abstract what they know into executable artifacts. SkillsBench quantifies the effect~\cite{li2026skillsbench}: across 84 expert-curated tasks, skills improved agent task completion rates by 16 to 23 percentage points (Claude Opus 4.5 rose from 22.0\% to 45.3\%). The baseline remains modest, but the direction is clear: AI competence measurably increases when the system assimilates externalized human expertise.

\subsection{From Professional to Personal Externalization}
\label{sec:personal}

The evidence above concerns professional expertise. In early 2026, users began extending the same skill paradigm to personal identity (Section~\ref{sec:intro}). Second Brain Starter~\cite{secondbrain2025} goes further: a Claude Code skill that builds a persistent, context-aware personal assistant. It stores user decisions, preferences, and context in structured markdown files, integrates with external services (Gmail, Slack, GitHub), and operates at configurable proactivity levels from passive observer to active partner. This architecture implements Coupled integration (Level 2): the system adapts to the user across sessions, and the user adapts workflows around the system's persistent memory.

Most current identity skills remain at the Static or Reactive level (0--1): persona descriptions without bidirectional adaptation or functional equivalence. Users are nonetheless spontaneously attempting to externalize individual behavioral patterns into executable digital artifacts. The trajectory from static encoding toward adaptive, bidirectional systems is the progression GCE predicts, and, crucially, the trajectory from external model \emph{of} a person to causally embedded component \emph{within} that person's cognitive architecture.

\subsection{Corporate Cognitive Distillation}
\label{sec:corporate}

The most commercially advanced form of cognitive externalization is emerging in enterprise settings, where organizations systematically distill employees' expertise into executable AI skills. A worker's conversational history, standard operating procedures, decision logs, and completed artifacts are extracted, structured into a skill artifact, and loaded into an AI agent that acts on the worker's behalf, responding to colleagues, making routine decisions, and handling domain-specific tasks in ways that recipients often cannot distinguish from the original person. This pipeline satisfies all three GCE criteria (Section~\ref{sec:framework}): the agent generalizes to novel situations, updates continuously from ongoing interactions while the worker adapts workflows around its presence, and operates inside the causal loop by sending emails, approving requests, and making decisions with real-world consequences that feed back to the worker.

Current deployments distill a snapshot (Level 0--1), but the commercial incentive is to distill continuously, accumulating behavioral data across a worker's full tenure. GCE predicts that with sufficient duration and data density, such systems will cross the Coupled threshold (Level 2) and approach Substitutive fidelity (Level 3) for well-defined professional functions. The same logic extends to any individual who grants an AI system persistent access to their behavioral outputs. How fast convergence proceeds is an empirical question GCE's predictions (Section~\ref{sec:predictions}) are designed to answer.

\subsection{Differential Externalization}

Not all cognitive functions externalize equally, and the NBIR hypothesis explains why. Functions with explicit behavioral signatures externalize more readily because NBIR is most plausible for them: their functionally relevant structure manifests in observable behavior, leaving no invisible residual that a coupled system would miss. At one end, decision-making, planning, preference expression, knowledge retrieval, and communication style all leave clear behavioral traces that AI systems can learn from. Problem-solving strategies, learning patterns, and attention allocation occupy a middle ground. At the other end, phenomenal experience, emotional qualia, self-awareness, and metacognition are hardest to capture because they may possess behaviorally invisible components, precisely the residuals whose existence would falsify NBIR for those domains. GCE therefore predicts differential rates: behaviorally manifest functions externalize first, producing partial but expanding coverage over time. The functions for which NBIR is least plausible define GCE's empirical boundary.

\subsection{From Individual to Collective Externalization}

Beyond differential rates across cognitive functions, externalization also scales from individuals to collectives. Infrastructure for multi-domain cognitive externalization has already reached population scale. OpenClaw, a self-hosted open-source AI assistant\footnote{\url{https://github.com/openclaw/openclaw}}, gained rapid adoption by providing persistent local memory, multi-domain integration across over fifty services, and privacy-preserving on-device processing~\cite{openclaw2026}. When many such human-agent dyads coexist, their interactions generate collective cognitive patterns that, under TAME's multiscale competency architecture, may constitute a new hierarchical level.

MiroFish~\cite{mirofish2026} illustrates a more radical form of collective externalization. Built on the OASIS framework~\cite{oasis2025}, it constructs parallel digital worlds populated by thousands of LLM-driven agents with independent personality, memory, and behavioral logic, externalizing not individual cognition but the \textit{collective cognitive dynamics} of social groups (opinion formation, coalition building, and preference contagion). The user can inject new variables and converse with individual agents, creating a meta-cognitive loop over externalized social cognition.

\section{Theoretical Foundations and Formal Framework}
\label{sec:framework}

GCE rests on three empirical premises: that cognitive behavior is learnable from observation, that it organizes hierarchically, and that distributed components can cohere into unified wholes, together with functionalism, the thesis that cognitive functions are individuated by their causal-functional roles rather than by substrate~\cite{putnam1967_psychological}. A system that plays the same causal-functional role as a biological cognitive process, with the same input-output profile, generalization patterns, and error structure, instantiates that function regardless of substrate. Combined with the NBIR hypothesis (Section~\ref{sec:intro}), this yields GCE's core claim: for cognitive functions whose behavioral output lies on a learnable manifold, a causally coupled system that reproduces the function's behavioral profile has externalized it. We first present evidence for each premise, then formalize three criteria for cognitive integration as measurable quantities.

\subsection{Behavioral Manifold Hypothesis}
\label{sec:bmh}

\citeauthor{fields2022_competency} characterize cognition as navigation through embedding spaces via error minimization~\cite{fields2022_competency}. This pattern appears across biological scales, from single cells navigating toward homeostatic setpoints~\cite{levin2019_computational} to cellular collectives maintaining target morphologies~\cite{levin2023_bioelectric} to population-level neural coordination~\cite{miller2024_cognition}. Direct empirical evidence confirms that the resulting activity occupies low-dimensional manifolds.~\citeauthor{fragoso2019_dimensionality} measured human behavioral activity data collected via smartphones and found a fractal dimension between 1.8 and 1.9, despite embedding in a seven-dimensional space~\cite{fragoso2019_dimensionality}. At the neural level,~\citeauthor{gallego2017_manifolds} showed that motor cortex population activity during movement is confined to a low-dimensional manifold shared across tasks~\cite{gallego2017_manifolds}.

Artificial systems converge on similar structures. Transformers learn relational embeddings~\cite{vaswani2017_attention}, and diffusion models navigate latent spaces through error-minimizing trajectories~\cite{ho2020_denoising}.~\citeauthor{huh2024_platonic} call this the ``Platonic representation hypothesis''~\cite{huh2024_platonic}: biological and artificial systems converge because the data-generating process has low intrinsic dimensionality. GPT-series models performed at high levels on Theory of Mind tasks without explicit training~\cite{kosinski2023_tom}, suggesting that even high-level social cognition lies on a learnable manifold, though this finding is contested, with adversarial evaluations showing failure on minor task perturbations~\cite{ullman2023_tom}, and GCE does not depend on it.

BMH entails a strong prediction: if everyday cognition occupies a low-dimensional manifold, then the manifold captures the functionally relevant structure of that cognition. We note that current dimensionality evidence comes from specific domains (smartphone activity, motor cortex); whether communication style, preference formation, and knowledge organization are equally low-dimensional remains an empirical question that must be answered domain by domain before GCE's predictions apply. The NBIR hypothesis sharpens BMH into GCE's central bet: the functions for which NBIR holds are the functions GCE can explain; those for which it fails, where a causally coupled system reproduces all behavioral outputs yet demonstrably fails to instantiate the cognitive function, define GCE's empirical boundary.

\subsection{Hierarchical Organization and Distributed Coherence}

If cognitive behavior is learnable, the next question is how learned components organize into larger wholes. Levin's Technological Approach to Mind Everywhere (TAME) framework demonstrates that cognitive capabilities emerge through hierarchical organization of competent subunits~\cite{levin2022_technological}. Biological systems instantiate nested competency hierarchies (molecular networks, cells, tissues, organs, organisms), where each level exhibits goal-directed behavior~\cite{levin2023_bioelectric} and cognition arises as an emergent property of population-level coordination~\cite{miller2024_cognition}. GCE extends this hierarchy by treating ambient AI systems as additional competent levels. The key principle TAME establishes is that higher-level cognition emerges from information-theoretic integration, not from any particular substrate~\cite{levin2019_computational}. If that principle holds, then adding digital subunits to the hierarchy is not a category error but a natural extension.

If cognitive processes distribute across substrates, what holds the system together? Sheaf theory formalizes conditions under which local information (from separate AI subsystems handling planning, communication, preferences) integrates consistently into global structures~\cite{fields2023_control}. The free energy principle provides a complementary account: coupled systems that mutually minimize prediction error form a joint Markov blanket~\cite{friston2010_free}. When users adapt to AI recommendations while AI models update from user feedback, both sides reduce mutual prediction error, forming a single functional unit.

\subsection{Bidirectional Adaptation}

The premises above establish that cognitive externalization is possible. The next question is when it actually happens. Not all AI systems qualify as cognitive extensions; users adapt to spell-checkers without those tools becoming cognitive extensions. To exclude such cases, we require three jointly necessary conditions. The AI must adapt its internal representations to learn decision structures specific to the individual user, exceeding a population-level baseline. The human must adapt cognitive patterns in ways that depend on the specific AI system: removing it produces measurable performance degradation beyond switching between equivalent tools. These adaptations must be mutually dependent and temporally coupled over a sustained period (we propose six months minimum).

We formalize this through mutual information:
\begin{equation}
    I(H_t; A_t) > \tau
\label{eq:mutual_info}
\end{equation}
Mere tool use shows low mutual information because the tool's internal state does not depend on the user's cognitive evolution. More deeply coupled systems differ:~\citeauthor{barke2023_copilot} report that Copilot users restructure their programming cognition around the system's capabilities~\cite{barke2023_copilot}, and~\citeauthor{jakesch2023_human} show that ChatGPT interaction shifts users' expressed opinions toward model outputs, persisting after the interaction ends~\cite{jakesch2023_human}. An important alternative explanation must be acknowledged: these findings may reflect anchoring bias or cognitive deskilling rather than genuine integration. GCE distinguishes externalization from deskilling empirically: in externalization, the coupled human-AI system performs \emph{at least as well} as the human alone did at baseline, because cognitive function has been redistributed rather than lost; in deskilling, both the human alone \emph{and} the coupled system degrade on novel challenges. Empirical tests must measure coupled-system performance, not just individual performance after removal. Neither study above measures all three conditions simultaneously; doing so is a priority for future empirical work.

\subsection{Functional Equivalence}

Functional equivalence requires more than output matching. A lookup table that maps every observed input to the correct output is behaviorally indistinguishable on training data but has no cognitive structure. We therefore require three properties. First, \textit{output fidelity}: AI-generated outputs must be indistinguishable from human-generated ones in blind evaluation across the relevant input domain. Second, \textit{generalization}: the system must produce appropriate outputs on novel inputs outside its training distribution, at rates comparable to the human's own consistency on novel inputs. A system that memorizes without generalizing is a record, not a cognitive extension. Third, \textit{structural correspondence}: the system's internal representations must exhibit relational structure that maps onto the target cognitive domain (e.g., preference embeddings that preserve the user's similarity judgments, not arbitrary encodings that happen to produce correct outputs).

We formalize functional equivalence through an instantiation measure. Let $\mathcal{C} = \{c_1, c_2, \ldots, c_n\}$ denote a finite set of observable cognitive functions (e.g., decision-making, temporal planning, communication style, preference expression). For each function $c_i$ and substrate $S$ (biological brain $H$, AI system $A$, or coupled pair), define:
\begin{equation}
    \phi(c_i, S) \in [0, 1]
\end{equation}
quantifying the degree to which $S$ realizes $c_i$, operationalized through behavioral tests. This operationalization is explicitly functionalist: $\phi$ measures behavioral-functional realization, and GCE's commitment (via NBIR) is that behavioral-functional realization is sufficient for cognitive function; no behaviorally invisible residual is needed beyond what $\phi$ captures. Since cognitive functions externalize gradually, we track overall progress through a time-dependent externalization ratio:
\begin{equation}
    E(t) = \frac{\sum_{i} w_i \cdot \phi(c_i, A_t)}{\sum_{i} w_i \cdot \phi(c_i, H_0)}
\label{eq:externalization}
\end{equation}
where $w_i$ weights the relative contribution of each cognitive function, $A_t$ denotes the state of the AI system at time $t$, and $H_0$ is the human baseline. $E(t) \in [0, 1]$ represents the overall degree of externalization. (We use $E$ rather than $\Phi$ to avoid confusion with IIT's intrinsic information measure $\Phi_{\text{IIT}}$~\cite{tononi2004_information}; GCE measures behavioral similarity, not integrated information.) The weights $w_i$ are not free parameters: they should be set empirically through the frequency and consequentiality of each cognitive function in the user's daily activity, measured via behavioral logs before the AI system is introduced. Sensitivity analysis across plausible weightings should accompany any reported $E(t)$ value. The functional equivalence threshold requires:
\begin{equation}
    \phi(c_i, A_t) \geq (1 - \delta) \cdot \phi(c_i, H_0) \quad \text{for all } c_i \in \mathcal{C}'
\label{eq:func_equiv}
\end{equation}
where $\delta$ is a domain-specific tolerance and $\mathcal{C}' \subseteq \mathcal{C}$ is a target subset.

Current systems approach output fidelity in narrow domains (Section~\ref{sec:evidence}), but generalization remains less well documented: most evaluations test on in-distribution data, and longitudinal studies of personalized AI performance on novel inputs are needed.

\subsection{Causal Coupling}

The third criterion requires that human and AI components be causally intertwined, not merely correlated. Causal coupling changes the ontological status of the other two criteria: a system that achieves bidirectional adaptation and functional equivalence \emph{without} causal coupling remains a high-fidelity external model. A system that achieves all three is no longer external to the cognitive process it tracks, and the ``model'' is inside the causal loop, and the map-territory distinction loses operational meaning~\cite{adams2001_bounds}.

Formally, for intervention $\text{do}(A_t := a')$:
\begin{equation}
    P(H_{t+1} \mid \text{do}(A_t := a')) \neq P(H_{t+1})
\label{eq:causal}
\end{equation}
This condition, testable through randomized perturbation experiments, rules out systems that merely correlate with human behavior. Causal coupling is a matter of degree: a search engine shapes information-seeking behavior, but the effect is generic rather than user-specific. GCE requires that the causal influence be \emph{personalized}: interventions on the AI's state must differentially affect \emph{this} user's cognitive outputs relative to other users, reflecting learned individual structure rather than generic tool effects. An integrated system should exhibit three observable signatures: correlated state changes, coordinated responses to novel inputs, and degraded performance when the components are decoupled.

To measure whether human and AI cognitive outputs are converging, we define a distance in a shared representation space. Let $\mathbf{h}_t$ and $\mathbf{a}_t$ denote vector representations of human and AI cognitive outputs at time $t$. Cognitive convergence holds when:
\begin{equation}
    d(\mathbf{h}_t, \mathbf{a}_t) < \epsilon \quad \text{for all } t > T
\label{eq:convergence}
\end{equation}
for some threshold $\epsilon$ and convergence time $T$, calibrated empirically by measuring inter-human variability on the same tasks.

\subsection{Integration Depth Taxonomy}

The criteria above define integration as a matter of degree. Table~\ref{tab:integration} organizes this continuum into five levels, with a key ontological boundary at Level~2. Static and Reactive systems (Levels 0--1, $E < 0.2$) constitute tool use: the system remains an external model \emph{of} the user's cognitive functions, with no causal embedding. At Level~2 (Coupled, $0.2 \leq E < 0.5$), bidirectional adaptation begins and the system enters the user's causal loop, marking the ontological transition from model to component. Substitutive and Integrated systems (Levels 3--4, $E \geq 0.5$) achieve functional equivalence across one or more cognitive domains while causally coupled; at these levels, the map-territory distinction has dissolved for the relevant functions. GCE predicts that current ambient systems are approaching the Substitutive level in specific domains.

\begin{table}[t]
\centering
\small
\begin{tabular}{@{}>{\raggedright\arraybackslash}p{1.9cm}>{\raggedright\arraybackslash}p{2.1cm}>{\raggedright\arraybackslash}p{4cm}@{}}
\toprule
\textbf{Level} & \textbf{Category} & \textbf{Example Systems} \\
\midrule
0: Static & Storage only & Notebook, file system \\
1: Reactive & Unidirectional learning & Autocomplete, spell-check \\
2: Coupled & Bidirectional adaptation & Copilot, personal AI assistants \\
3: Substitutive & Functional equivalence & Predictive scheduling, style-matched writing \\
4: Integrated & Multi-domain coherence & Cross-domain personal AI (conjectural) \\
\bottomrule
\end{tabular}
\caption{Integration depth taxonomy. Levels 0--1 (Static, Reactive) constitute tool use. Levels 2--4 (Coupled, Substitutive, Integrated) represent increasing degrees of cognitive integration.}
\label{tab:integration}
\end{table}

\subsection{Scope and Boundaries}
\label{sec:scope}

GCE's predictions (Section~\ref{sec:predictions}) are strongest for domains with demonstrated low dimensionality (temporal planning, preference expression, communication style) and progressively more speculative for higher-level functions. Population-level low dimensionality also does not guarantee low dimensionality for individual variation. If individual cognitive manifolds prove high-dimensional, GCE would hold only for shared low-dimensional components, not for the idiosyncratic aspects that most constitute personal identity. Additionally, output convergence alone does not establish cognitive alignment: both systems could converge toward a shared cultural mean, or the human could anchor on AI outputs~\cite{jakesch2023_human}. Empirical tests should therefore require dyad-specific convergence that exceeds non-personalized baselines.

A natural objection holds that cognitive processes may not be computable, and therefore cannot be replicated digitally. But learnability and computability are distinct: a function class can be learnable even when individual target functions are not computable~\cite{gold1967_language, valiant1984_learnable}. Computability requires exact evaluation on every input, while learnability requires only asymptotic behavioral indistinguishability under the relevant distribution. GCE exploits this gap. Ambient systems need not compute the internal cognitive processes they externalize, only learn the behavioral manifold those processes generate. This accommodates non-computationalist views: even if the generating process is quantum or non-algorithmic, its behavioral output can still be learned from observation. The practical bottleneck for mind uploading was neural observation. Ambient intelligence sidesteps it by accumulating behavioral data from the outside.

A second objection holds that ambient AI creates ``guardians'' or models \textit{of} persons rather than instantiations \textit{of} their cognitive processes, arguing that a highly accurate weather model does not externalize the atmosphere. This is GCE's strongest challenge. Our response: the weather-model analogy fails because of causal embedding (Section~\ref{sec:framework}). A weather model sits outside the atmosphere; an AI system in a continuous feedback loop with a user is inside the cognitive process (Equation~\ref{eq:causal}). The objection implicitly asserts a behaviorally invisible residual present in ``real'' cognition but absent from the model, precisely what NBIR denies. If such a residual is behaviorally invisible by definition, it cannot play a functional role~\cite{putnam1967_psychological, block1980_troubles}; if it \textit{is} functionally relevant, it must manifest behaviorally, and the coupled system will eventually learn it. When all three criteria are jointly satisfied, the burden shifts to the critic: specify an observable, operationalizable property that distinguishes the coupled system from a genuine cognitive extension~\cite{adams2001_bounds}. If none can be specified, the distinction is empirically empty.

A related objection holds that trivial systems also satisfy GCE's criteria: a thermostat in a feedback loop with a house adapts bidirectionally (heater changes room temperature, which changes thermostat state), achieves a form of functional equivalence (room stays at target), and is causally coupled. Yet no one claims the thermostat externalizes the homeowner's temperature-regulation cognition. The response is that the three criteria are \textit{jointly} necessary, and the thermostat fails functional equivalence as GCE defines it: it does not generalize to novel inputs outside its operating range, does not exhibit structural correspondence with the homeowner's thermal preferences across diverse contexts, and its internal state has no relational structure that maps onto the target cognitive domain. A system that memorizes a single setpoint without generalizing is a controller, not a cognitive extension. GCE's functional equivalence criterion (Section~\ref{sec:framework}) requires output fidelity, generalization to novel inputs, and structural correspondence, conditions that exclude trivial control loops while admitting systems of genuine cognitive complexity.

Biological identity already involves gradual change: neurons are replaced, memories transform, personality evolves. What preserves identity is pattern continuity, not substrate permanence. GCE extends this principle to biological-digital substrates without a discontinuous transfer event. Unlike the Ship of Theseus, GCE involves component \textit{addition}, as the biological substrate remains intact, though functional substitution may produce de facto replacement over time, a risk the autonomy preservation principle (Section~\ref{sec:predictions}) addresses.

GCE does not claim qualia transfer; readers who reject functionalism will reject GCE on principled grounds~\cite{block1980_troubles}. Against Searle's Chinese Room~\cite{searle1980_minds}: unlike Searle's isolated room, ambient systems are causally embedded in feedback loops with environments and users, satisfying the ``robot reply'' that Searle acknowledged as the strongest counterargument. Against Penrose: even if cognition is non-algorithmic, its behavioral output occupies a learnable manifold~\cite{gold1967_language}, and AI systems now perform tasks Penrose claimed require non-algorithmic insight~\cite{trinh2024_alphageometry}.

If a system learns a person's cognitive manifold with high fidelity, are there now two instances of that cognition? Under GCE, the externalized system is a \textit{coupled extension}, not a copy. If severed, it becomes an autonomous replica that can instantiate the person's cognitive functions. Whether this constitutes identity persistence depends on $E(t)$ at severance, and the framework converts a metaphysical debate into a quantitative empirical question.

Finally, GCE is not reducible to ordinary cognitive offloading~\cite{risko2016_cognitive}, which is unidirectional: a person delegates to an external resource that does not adapt to the person's cognitive structure.~\citeauthor{heersmink2015_dimensions} taxonomize cognitive integration along dimensions including information flow direction, from monocausal through bicausal to reciprocal continuous causation~\cite{heersmink2015_dimensions}. GCE's three criteria demarcate the boundary where offloading transitions into integration at the high end of this spectrum.

\section{Research Agenda}
\label{sec:predictions}

A framework's value is partly measured by the experiments it enables. Each formal quantity in Section~\ref{sec:framework} maps to a specific empirical prediction, with thresholds constrained by the theory rather than chosen as free parameters. Together, these form a protocol that future longitudinal studies can directly implement. Consistent with the caveats in Section~\ref{sec:scope}, our strongest predictions concern domains where BMH has empirical support (temporal planning, preference expression, communication style). Predictions for higher-level functions (reasoning, creativity) are more speculative and should be treated as hypotheses requiring independent validation of low dimensionality before the convergence thresholds apply.

\subsection{Behavioral Convergence}

AI systems trained on extended behavioral observation will predict user decisions with accuracy approaching human self-prediction baselines (Prediction 1). The threshold is not arbitrary: functional equivalence (Equation~\ref{eq:func_equiv}) requires that $\phi(c_{\text{decision}}, A_t) \geq (1 - \delta) \cdot \phi(c_{\text{decision}}, H_0)$, where $\delta$ equals the inter-subject coefficient of variation in self-prediction accuracy. For a concrete example: if humans predict their own next-day scheduling decisions with 75\% accuracy ($\pm$8\% across individuals), the AI must reach 69\% to satisfy the criterion. As documented in Section~\ref{sec:evidence}, recommendation systems already approach this threshold in the preference domain~\cite{netflix2009_prize}.

AI-generated communications will become indistinguishable from human-generated ones (Prediction 2). The convergence metric (Equation~\ref{eq:convergence}) sets the threshold: the distance $d(\mathbf{h}_t, \mathbf{a}_t)$ must fall within the distribution of inter-human distances on the same task. In practice, this means judge accuracy in a forced-choice discrimination task falls to chance level (50\%) or within the range observed for human-vs-human discrimination. Evidence from Section~\ref{sec:evidence} suggests current writing assistants already approach this threshold~\cite{clark2021_human_eval, chen2019_gmail}.

\subsection{Bidirectional Adaptation}

Long-term AI users will exhibit measurable cognitive pattern changes reflecting AI characteristics (Prediction 3). The threshold derives from the mutual information criterion (Equation~\ref{eq:mutual_info}): $I(H_t; A_t)$ must exceed $\tau$, calibrated against mutual information between human collaborators performing the same task. If human-AI mutual information remains below human-human levels despite extended interaction, GCE's coupling claim fails.

Human-AI systems will exhibit correlated cognitive changes: model updates will correlate with user behavioral changes and vice versa (Prediction 4), as already documented in recommendation system co-evolutionary dynamics~\cite{pariser2011_filter}.

AI systems will predict user physiological states (stress, fatigue, emotional valence) with accuracy approaching human observer baselines (Prediction 5). The threshold follows the same logic as Prediction 1: accuracy within $\delta$ of human observer accuracy, where $\delta$ is the inter-observer coefficient of variation. This prediction is the most speculative of the five: it remains untested and depends on the assumption that physiological states produce consistent behavioral signatures, which is contested in affective computing research. We include it because it is directly derivable from GCE's logic: if ambient systems learn behavioral manifolds with sufficient fidelity, states that reliably modulate behavior should become predictable. However, it requires independent validation that the relevant physiological-behavioral mappings are sufficiently regular.

\subsection{Falsification Conditions}

GCE would be falsified if any of the following held: (1) AI prediction accuracy plateaus for more than two years despite order-of-magnitude increases in interaction data; (2) no measurable human adaptation occurs, indicating the relationship is unidirectional; (3) AI-generated content remains fundamentally distinguishable from human content regardless of training duration; (4) model and user behavioral changes show no correlation; or (5) a domain is identified where a causally coupled system reproduces all behavioral outputs with high fidelity yet demonstrably fails to instantiate the cognitive function, constituting direct falsification of NBIR for that domain. Condition (5) could be operationalized through downstream task transfer: if the coupled system matches the user's outputs on trained tasks but systematically fails on structurally related tasks that the user handles effortlessly, this would demonstrate that behavioral surface matching does not capture the underlying cognitive competence, constituting evidence for a functionally relevant residual. Condition (5) is the most theoretically consequential: it targets GCE's central assumption. All five are testable with current methodologies and specific enough to be confirmed or refuted within two to three years of longitudinal study.

\subsection{Design Implications}

The predictions and falsification conditions above define what to measure. The next question is what to build. Five design principles follow. First, \textit{longitudinal learning}: systems should accumulate information across extended periods rather than resetting between sessions~\cite{langchain2025_ambient}. Second, \textit{multi-domain integration}: combining cognitive domains increases overall $E(t)$. Third, \textit{bidirectional interfaces}: systems that lack bidirectional interfaces remain external models \emph{of} the user regardless of prediction accuracy; bidirectionality is what transforms the system from model to causally embedded component. Fourth, \textit{privacy-preserving personalization}: addressing the tension between deep behavioral data and privacy through federated learning and on-device processing. Fifth, \textit{autonomy preservation}: the same coupling that enables integration creates dependency risks; GPS users show degraded spatial memory from habitual offloading~\cite{dahmani2020_gps}, and coupled systems may amplify rather than represent preferences~\cite{pariser2011_filter, parasuraman2010_complacency}. Design must include decoupling safeguards: periodic assessments of independent performance, transparency about system influence, and graceful degradation paths.

\section{Conclusion}

GCE provides a falsifiable framework for understanding how ambient AI systems transition from modeling cognitive functions to constituting part of users' cognitive architectures. Evidence across scheduling, communication, preference, and knowledge domains shows that the preconditions are already observable. The framework's three criteria (bidirectional adaptation, functional equivalence, causal coupling), grounded in the NBIR hypothesis, generate five testable predictions with theory-constrained thresholds. The central question is whether causally coupled systems cross a threshold from modeling to constituting. To resist GCE's affirmative answer, one must either reject functionalism or identify a specific, operationalizable residual that a system satisfying all three criteria still lacks. If longitudinal studies fail to confirm the predicted convergence dynamics, or if behaviorally invisible residuals are demonstrated, GCE fails with them.

\bibliography{references}

\end{document}